\documentclass[letterpaper, 10 pt, conference]{ieeeconf}  

\IEEEoverridecommandlockouts                              

\usepackage{subcaption}
\usepackage{ragged2e, hyperref}
\usepackage{balance}
\usepackage{placeins, float}

\usepackage{subcaption}

\makeatletter
\let\NAT@parse\undefined
\makeatother

\usepackage{graphicx}
\usepackage{microtype}
\usepackage{relsize}
\usepackage{url}

\title{\LARGE \bf
``You've got a friend in me'': Co-Designing a Peer Social Robot for Young Newcomers' Language and Cultural Learning
}

\author{Neil Fernandes$^{1}$, Cheng Tang$^{1}$, Tehniyat Shahbaz$^{2}$, Alex Hauschildt$^{2}$, \\ Emily Davies-Robinson$^{2}$,
Yue Hu$^{1}$ and Kerstin Dautenhahn$^{1}$
\thanks{This research was funded, in part,
by the Canada 150 Research Chair Programme. This work has been submitted to the IEEE for possible publication. Copyright may be transferred without notice, after which this version may no longer be accessible}
\thanks{$^{1}$University of Waterloo, Waterloo, Canada; Emails:\{neil.fernandes, cheng.tang, yue.hu, kerstin.dautenhahn\}@uwaterloo.ca}%
\thanks{$^{2}$United for Literacy, Toronto, Ontario, Canada; Emails: tehniyat.shahbaz@gmail.com, ahauschildt@unitedforliteracy.ca, edrobinson@unitedforliteracy.ca}
}

\begin{document}

\maketitle
 \thispagestyle{empty}
\pagestyle{empty}

\begin{abstract}
Community literacy programs supporting young newcomer children in Canada face limited staffing and scarce one-to-one time, which constrains personalized English and cultural learning support. This paper reports on a co-design study with United for Literacy tutors that informed \textit{Maple}, a table-top, peer-like Socially Assistive Robot designed as a practice partner within tutor-mediated sessions. From shadowing and co-design interviews, we derived newcomer-specific requirements and added them in an integrated prototype that uses short story-based activities, multi-modal scaffolding (speech, facial feedback, gesture), and embedded quizzes that support attention while producing tutor-actionable formative signals. We contribute system design implications for tutor-in-the-loop SARs supporting language socialization in community settings and outline directions for child-centered evaluation in authentic programs.
\end{abstract}


\section{Introduction}
International migration is increasingly reshaping education and community support systems across the world. The global number of international migrants reached 304 million in 2024, and OECD countries recorded a record 6.5 million new permanent-type immigrants in 2023 \cite{undesa2024migrantstock}. These demographic shifts have direct consequences for local institutions that support newcomer integration, including community-based literacy programs that complement formal schooling and provide families with accessible language practice and guidance  \cite{anderson2015intersubjective}. Young newcomer children from immigrant and refugee backgrounds often learn English while simultaneously adapting to unfamiliar life, e.g.\ new school routines, institutional expectations, and broader socio-cultural norms in their new environment \cite{shahbazi2020breaking}. In community-based literacy programs, these challenges are compounded by practical constraints such as limited staffing, mixed-age and mixed-proficiency groups, and limited opportunities for sustained one-to-one practice \cite{OntarioMoE2008}. Tutors therefore have to balance instruction, emotional support, and attention management under tight time and resource limits. In this paper, we use the term \textit{newcomer children} to refer to young children from immigrant or refugee backgrounds who have migrated to Canada within the past five years and are in the process of learning English as a second language (L2), adopting prior definitions used in Canadian education research \cite{IRCC2018}.

\begin{figure}[t]
    \centering
    \includegraphics[width=1
    \linewidth]{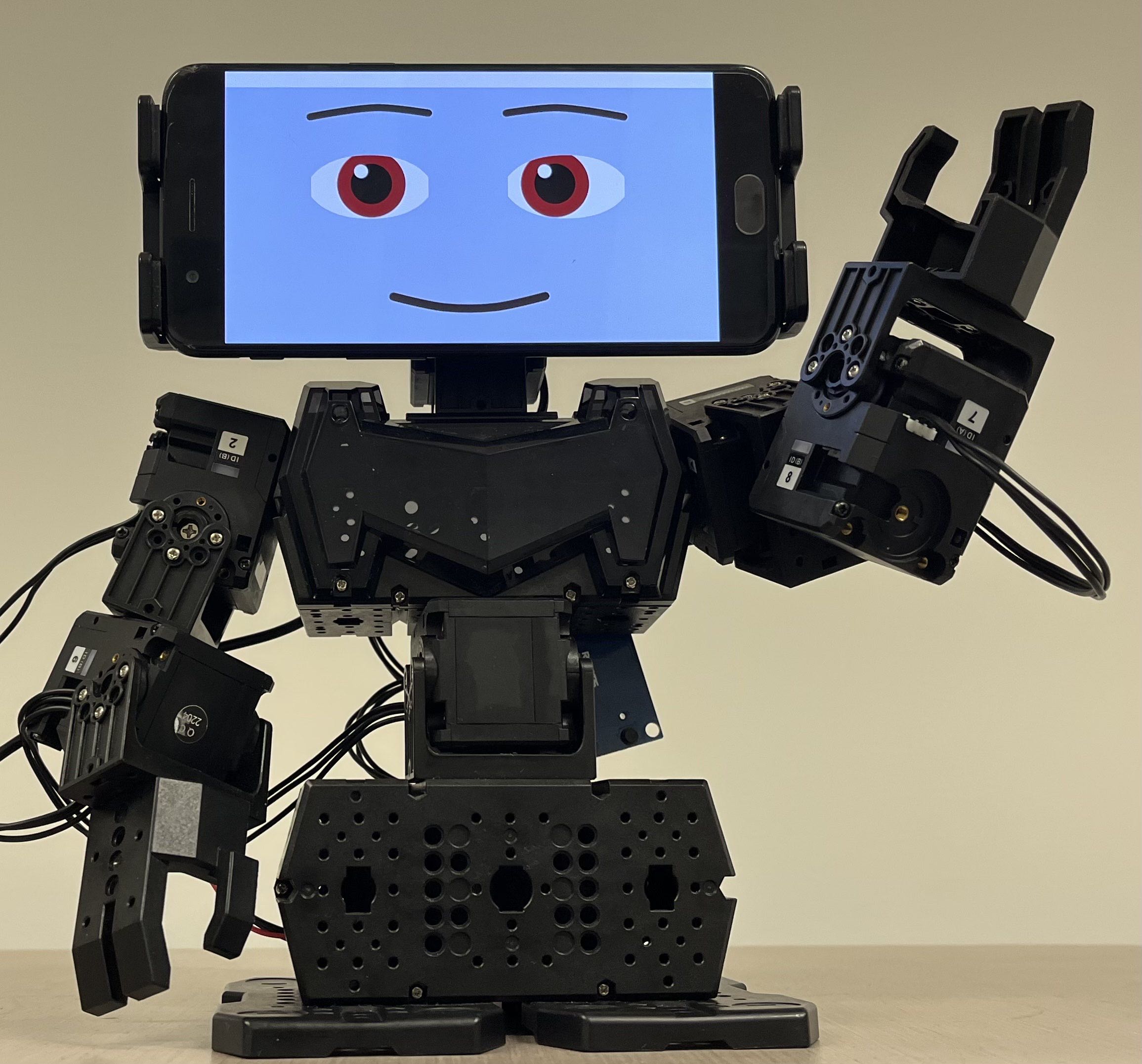}
    \caption{The \textit{Maple} robot prototype \cite{fernandes2026codesigning} (Licensed under CC BY-NC-ND 4.0).}
    \label{maple}
\end{figure}
\begin{figure*}[!h]
    \centering
    \includegraphics[width=1\linewidth]{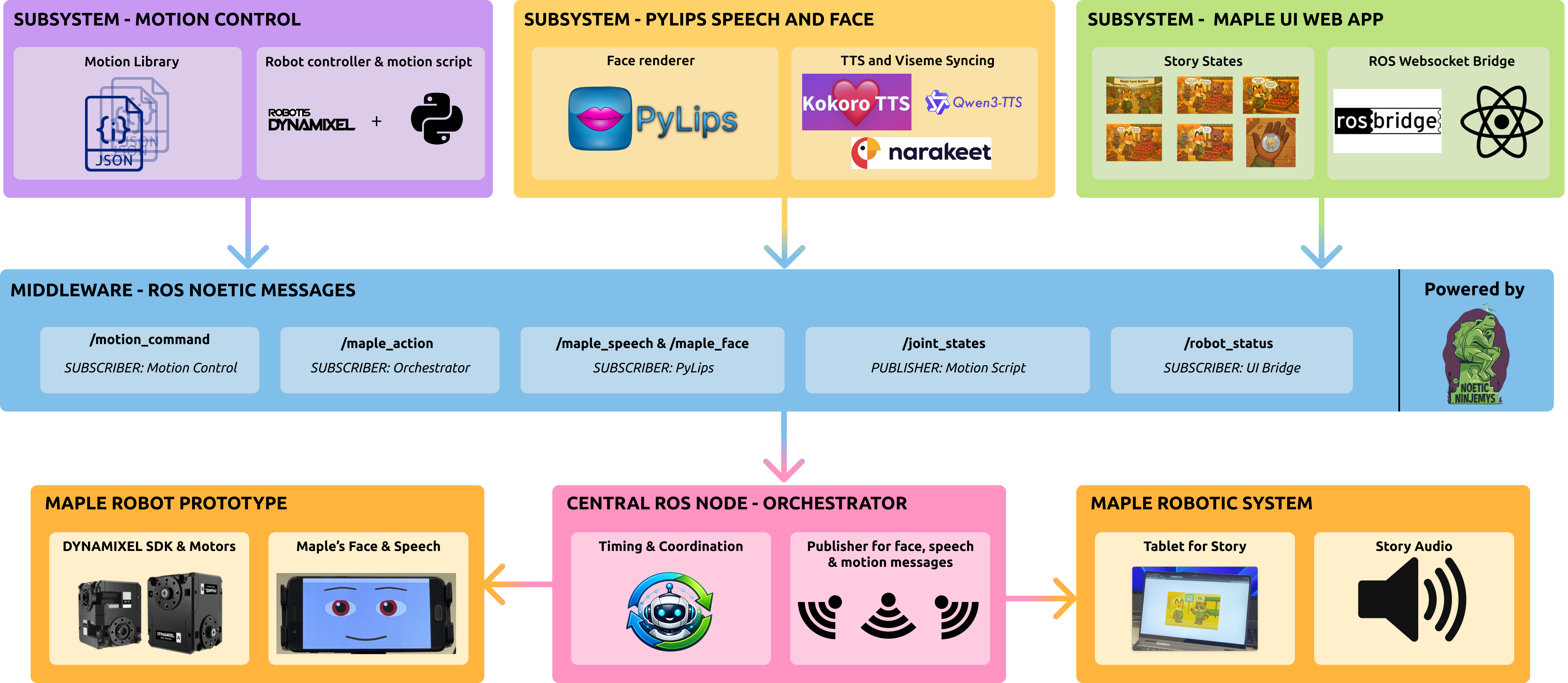}
    \caption{System Architecture of the \textit{Maple} robotic system and robot prototype.}
    \label{systemoverview}
\end{figure*}

Socially Assistive Robots (SARs) are increasingly explored as supports for language learning because they can make practice feel more engaging and socially meaningful. Prior work suggests that, even when long-term learning gains are mixed, robots can improve learners' motivation and confidence \cite{randall2019survey}. Compared to screen-based alternatives, social robots have also been associated with higher gains in child second-language practice, indicating that interaction framing and social presence can matter for learning experiences \cite{konijn2022social}. Physical embodiment is often argued to contribute to this effect, as robots may be perceived as more affiliative, socially present \cite{heerink2009social}, and socially engaging than disembodied systems such as virtual agents displayed laptops \cite{li2015benefit}. Repeated child-robot interaction has been shown to support sustained engagement in activities such as storytelling \cite{kory2019long}. These considerations become especially easy to focus on in newcomer settings. Learners are simultaneously developing an additional language and learning how to participate in new societal and institutional routines and cultural expectations \cite{OntarioMoE2008}. Anxiety and fear of negative evaluation can further inhibit willingness to speak \cite{horwitz1986foreign}, and concerns around cultural safety and trust shape what kinds of interactions are appropriate \cite{brown2020newcomer}. As a result, SARs may be most useful in community programs when designed not as stand-alone instructors, but as peer practice partners that provide repeatable, low-pressure interaction while keeping tutors in the loop. This framing aligns with emerging work on culturally responsive design \cite{louie2022designing} and culturally sensitive mediation \cite{yang2024towards}. Our prior work \cite{fernandes2026codesigning} addressed this gap by deriving five preliminary expert-grounded guidelines for SARs that support English learning and Canadian socio-cultural norms for newcomer children in community settings, which are: (i) provision of multi-modal and multilingual scaffolding for language barriers, (ii) usage of story-based activities to support attention, (iii) learning a second language through cultural orientation, (iv) embedding formal assessments into playful interactions and (v) supporting one-on-one attention within a triadic interaction. In this paper, we use scaffolding to mean temporary support that helps learners understand, participate in, and complete tasks that they may not yet be able to manage independently.

Building on these insights, we focus on language learning together with cultural learning in terms of \textit{language socialization}, where acquiring language and acquiring culturally grounded ways of participating in social life are interdependent \cite{schieffelin1986language}. We partnered with United for Literacy (UFL), a national charitable literacy organization serving children, youth, and adults in urban, rural, and remote communities across Canada, offers direct insight into learners’ needs and the constraints encountered in tutoring practice, to co-design \textit{Maple} (see Figure \ref{maple}), a table-top humanoid robot intended to function as a peer practice partner within tutor-mediated community sessions. Rather than positioning the robot as an autonomous instructor, our approach emphasizes predictable interaction and clearly identified situations within the system where tutors can step in for clarification, repair, and culturally sensitive guidance. This paper builds on our previous work on \textit{Maple} \cite{yang2024towards} by presenting a co-designed robot prototype that applies expert-identified requirements for supporting newcomer children in community literacy programs. This is done through short story-based activities and embedded quizzes. We also synthesize design implications for tutor-mediated SARs that support language learning and socialization goals in real community settings.


\section{Research Questions}
While the literature shows a significant potential of SARs for second language learning, tying language learning with the acquisition of new cultural norms - a concept encapsulated in \textit{Language Socialization} \cite{schieffelin1986language} - remains underexplored in SARs. In light of this, this paper aims to answer the following research questions:

\begin{enumerate}
    \item [RQ1:] How do L2 experts view the potential use of a social peer robot for L2 and cultural orientation?
    \item [RQ2:] What recommendations do L2 experts have for designing such a system in terms of interaction design and lesson content?

\end{enumerate}

In this paper, we define ``Cultural Orientation'' and ``Cultural Learning'' as the acquisition of everyday Canadian social norms, pragmatic competence, and community navigation skills that are typically learned alongside (or outside) English in community settlement programs. L2 experts refers to tutors with expertise in second-language learning and teaching, particularly in supporting learners who are acquiring English as an additional language.


\section{System Implementation}

\subsection{Maple Robotic System}
\textit{Maple} is a table-top humanoid robot designed to support interactive, story-based learning activities with young newcomer children. The system is built to deliver lesson content through verbal and non-verbal cues, while remaining lightweight and user friendly to deploy in small-group or classroom-like settings. The robot is built from the ROBOTIS Engineering Kit 1\footnote{\url{https://en.robotis.com/model/page.php?co_id=prd_engineerkit}}, evaluated from previous work \cite{yang2024towards}. The system can be powered either through a DC barrel connector, or a LiPo. From a system perspective (as seen in Figure \ref{systemoverview}),  \textit{Maple} is organized between (i) low-level device control, (ii) interaction content and operator-facing control, and (iii) higher-level behavior coordination. This separation allows each capability to be developed and tested independently while still producing integrated robot behaviors before (or during)  the activity. At runtime, the robot executes behaviors (for example, a gesture paired with a spoken line and an appropriate facial response) in response to interaction events, enabling consistent lesson flow and predictable tutor oversight. The robot can be coupled with a tablet on the side to display lesson content. These details are expanded in the next few sections.

\subsection{Control System Architecture}
At the hardware layer, \textit{Maple}’s DYNAMIXEL motors communicate over a shared serial bus. A motion controller initializes the bus, enables motor torque, and issues coordinated commands using synchronous group operations so multiple joints move together. Motions are triggered by middleware events and are specified in external motion files that define the relevant motors, a sequence of target joint poses, and timing parameters (for example brief holds between poses). During execution, the controller steps through each pose and sends a single grouped command to move all involved motors simultaneously, preserving the intended pacing of gestures. At shutdown, the controller safely disables torque and closes the serial interface. Defining gestures externally decouples motion authoring from control logic, making it straightforward to add or tune gestures without modifying the controller code.

\subsection{UI System Integration}
A web-based interface, implemented in React, provides users with an opportunity to interact with lesson content or videos. The interface establishes a websocket connection to a Robotic Operating System (ROS) bridge and registers handlers for connection status and errors. During interaction, the UI publishes action messages that can trigger robot behaviors (such as a gesture, a spoken line, or a facial response). Interaction content is driven by a real-time architecture called RoboSync \cite{tang2023robosync}, that defines a sequence of states for storytelling and quiz activities. Each state can display on-screen media (for example an image), audio playback, instructional or narrative text, and an associated robot behaviour to execute in sync. State transitions can be time-based (advance after a specified duration) or input-based (advance after the participant selects an answer). We also implemented quiz states, in case the lesson content required short quizzes between story elements. The configuration includes multiple-choice options and a correct answer for each quiz (in order to elicit a positive or negative reaction to the answer).

\subsection{Facial Animation and Speech Layer}
PyLips from Denler et. al. \cite{dennler2024pylips} is used to render \textit{Maple}'s face on the smartphone display as a browser-based interface, providing animated eyes, mouth motion, and configurable expressions. In our system, facial behaviors serve two roles: (i) they provide socially meaningful feedback during tutoring tasks and (ii) they improve perceived embodiment by coupling speech output with synchronized mouth visemes. To support the customization of facial features, Action Unit parameters and Expression presets can be changed at runtime, allowing the face style and affective cues to be adapted across scenarios. This customization also allowed the research team to adjust the default facial features in order to make it suitable for our application. The speech used for the purposes of the prototype were made using an open source Text-to-Speech (TTS) model called Kokoro-TTS\footnote{\url{https://huggingface.co/spaces/hexgrad/Kokoro-TTS}}.

\subsection{Node Orchestrator and Synchronization}
A dedicated orchestration layer coordinates the independently implemented subsystems via ROS Nodes into a behavior stream. The orchestrator receives high-level action messages from the controller and UI interprets them as a structured command that may include a gesture request, a speech request, and a facial expression or appearance update. It then schedules these elements according to the scenario’s configuration file stored in the UI. For example, speech may be initiated before a gesture to establish context, or facial feedback may be triggered in parallel with a short affirmative gesture. By centralizing timing decisions in the orchestrator, the system avoids tightly coupling the motion controller to the face and audio components.

\section{Methods}
\subsection{Co-design Participants}
Our co-design team comprised six participants: three second language learning (L2) Expert Tutors (referred to in the paper as ``L2 experts'') from United for Literacy (UFL), a national organization that delivers community literacy programs across Canada\footnote{\url{https://www.unitedforliteracy.ca/Programs}}, and three Human-Robot Interaction (HRI) researchers (two faculty members and one graduate researcher). One L2 expert was a regional manager overseeing UFL programs in the Region of Waterloo, with seven years of experience, one L2 expert was a program support staff in the same region, with two and a half years of experience with UFL as both staff and volunteer and another L2 expert was a lead who was recently on-boarded, with 10 years of experience working as a High School teacher and 12 years of experience working with refugee and newcomers. All L2 experts involved had experience supporting newcomer children and families through the YMCA of Three Rivers Family Language and Learning Program\footnote{\url{https://www.ymcathreerivers.ca/immigrant-services-programs}}.

\subsection{Co-design Methodology and Sessions}
We employed a co-design approach \cite{garell2024codesign} aligned with a ``Design Thinking'' methodology \cite{brown2008design}, leveraging expertise from the L2 experts to align technology with user needs. Across the different stages, the co-designers actively contributed to generating and refining solutions grounded in the day-to-day constraints of community language learning for newcomer children. Figure \ref{fig:maple_timeline} summarizes the timeline of our co-design activities. These activities comprised of:



\begin{figure*}
    \centering
    \includegraphics[width=1\linewidth]{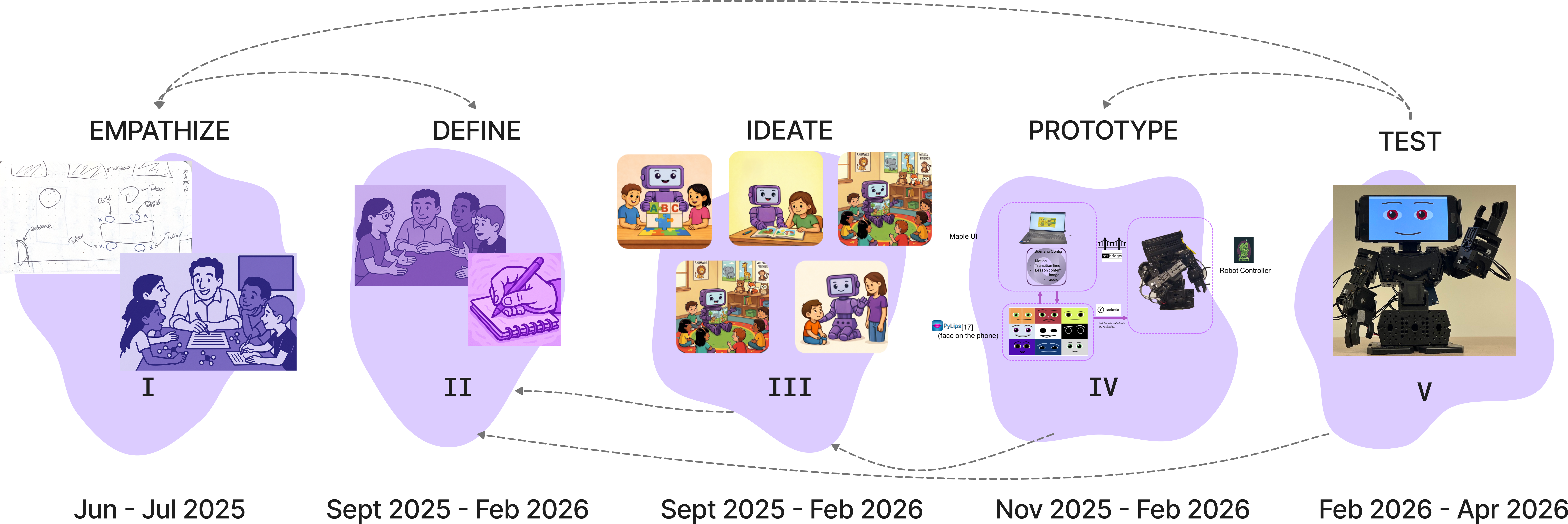}
\caption{Timeline of our co-design process with L2 and HRI experts, structured around Design Thinking: \textbf{I} (Empathize): This stage included Shadowing Sessions and Observational Analysis. \textbf{II} (Define): Stage included Transcript analysis, synthesis of challenges, and identification of user needs. \textbf{III} (Ideate): Included robot role definition, generation of design guidelines, and co-design of story based lesson activities. \textbf{IV} (Prototype): Included implementation and iterative refinement of the interactive system. \textbf{V} (Test): Included (and will include) empirical evaluation through tutor playtesting, interviews, and child case studies. }
    \label{fig:maple_timeline}
\end{figure*}

\subsubsection{Shadowing Sessions} \label{shadowsesh}
Shadowing is a qualitative field method used to capture real-time information about individuals' interactions and routines in natural settings. Czarniawska-Joerges characterizes it as \textit{``fieldwork on the move''} \cite{czarniawska2007shadowing}. During the empathize stage, we conducted shadowing sessions within ongoing YMCA language learning classes to understand how tutors delivered content, managed attention, and responded to language barriers in situ. Shadowing was conducted by one graduate HRI researcher, who took anonymous observational field notes focused on tutor-child interaction patterns, activity flow, materials used, and contextual constraints. These sessions served as an entry point into the problem space, providing grounded context for subsequent co-design discussions. In particular, the observational notes helped identify practical constraints and features specific to community program delivery. They also helped in brainstorming certain ways of including \textit{Maple} in the current workflow of language programs offered at UFL.

\subsubsection{Expert Group Interviews}
We conducted four semi-structured interview sessions with UFL tutors. Each session lasted approximately 60 minutes and was held online. The first four sessions were conducted as group interviews with two UFL tutors and the three HRI researchers. The fifth session was conducted as an individual interview between one UFL tutor and one HRI researcher to allow deeper reflection on the discussions that emerged in the group sessions. The first session followed a two-part structure. First, we asked tutors about everyday practice and the challenges present in the community language learning program context. Second, tutors were shown a short demonstration video of the \textit{Maple} prototype, highlighting initial story content and associated gestures. Following this video demo, the co-design team brainstormed possible activities grounded in the robot's demonstrated capabilities. The team also brainstormed possible guidelines that could address some of the issues that many tutors and learners face in the educational context. The second session focused on how \textit{Maple} could be integrated into existing program activities, emphasizing adaptation to current routines rather than replacement. The third session focused on translating the emerging concepts into concrete study plans, including how a socially assistive robot could fit within a pre-existing community workshop and how to assess its practical value in real program settings. The fourth interview mirrored the first session's structure and used the same video stimulus to validate and extend prior ideas from an individual perspective.



\section{Results and Discussion}

\subsection{Guidelines and System Outcomes}
This section translates the expert-grounded guidelines from our prior work \cite{fernandes2026codesigning} into concrete system outcomes, and describes how each outcome is implemented in the current \textit{Maple} prototype.

\subsubsection{Multimodal scaffolding and story-based activities (Guidelines 1 and 2)}
To support multimodal scaffolding during story-based learning, we integrated \textit{Maple} with an interface that allows tutors to select on-screen visuals with the robot's spoken narration and gestures. Tutors select story scenes from a scene library (through the scenario configuration files), and the system synchronizes (i) the displayed illustration, (ii) the narrated utterance, and (iii) corresponding robot gestures. The story unfolds through a progression style similar to one used in a 'visual novel' games (e.g. Ace Attorney \footnote{\url{https://www.ace-attorney.com}}), where dialogue and scene changes are presented sequentially to support immersion and provide contextual grounding for vocabulary learning \cite{camingue2021visual}. In the current prototype, scenes are English-only so that beginner-level vocabulary remains easy to notice and focus on. Target words are drawn from the Dolch \cite{johns1970dolch} word list, representing high-frequency vocabulary that is common in early reading instruction. This choice was also taken into account from the observations seen at the Shadowing Sessions (see Section \ref{shadowsesh}). In addition to these observations, co-design discussions focused on learners aged 5 to 10 years, which aligns with early elementary schooling, and with findings that earlier second-language exposure is associated with stronger long-term outcomes than later starts \cite{snow1978criticalperiod}. Drawing on \textit{Social Learning Theory} \cite{bandura1977social} (embedded within \textit{Language Socialization}) and the \textit{Audio Lingual Method} \cite{zillo1973birth} (also seen in practice during the Shadowing Sessions), the system schedules repeated exposures to target words at planned points across each story so that vocabulary is revisited through consistent auditory (when the robot repeats the word three times and visual cues (from the story, which is displayed when the robot repeats the words). These planned points are also coupled with optional deictic communication, where the robot points at the screen while repeating the target words. The exposure of these words are done in the style of incidental learning \cite{hulstijn2001intentional}.

For prototyping, we expanded the story set in collaboration with the co-designers. Stories are presented as a simple children's storybook featuring Canadian animal protagonists, rather than culturally specific depictions of human children. This choice was made since character identity cues can shape narrative engagement and perceived self-relevance, and children's learning and recall from stories can differ as a function of the protagonists' race and related demographic cues \cite{dore2022effect}. Related work in narrative media similarly suggests that engagement and wishful identification can vary with racial presentation and ambiguity, including in animated and fantasy-based designs \cite{lu2024effect}. We therefore adopt animal protagonists as an inclusion-oriented design decision to reduce the risk of positioning any single cultural identity as the default within a diverse newcomer population.


\subsubsection{Learn a second language through cultural orientation (Guideline 3)}
To align language learning with cultural orientation, the story content was designed to reflect principles of \textit{Language Socialization }\cite{schieffelin1986language}. In practice, this means that linguistic targets are embedded within everyday social situations, rather than introduced as isolated vocabulary items. Animal protagonists also support a neutral introduction to local practices without anchoring the narrative in any single culturally specific human depiction \cite{pente2009hidden}. In this framing, the animals function as socially neutral guides through common scenarios, while preserving space for children's own identities and family traditions \cite{shaw2001family}. At the same time, we treat these motifs as cultural resources rather than universal traits, consistent with critiques that wilderness iconography can normalize particular versions of nationhood and belonging \cite{pente2009hidden}. This approach also aligns with Canada’s long-standing \textit{``mosaic''} framing of plural identities, where national belonging is often articulated through shared civic narratives rather than a single cultural default \cite{gibbon1938mosaic}.

\subsubsection{Embedding formal assessment into playful actions (Guideline 4)}
Experts emphasized that tutors need actionable information about learners' current language knowledge to personalize instruction, while worksheet-like assessment formats can be demotivating in community literacy contexts \cite{fernandes2026codesigning}. To address this, we embedded short quizzes within the story flow so that assessment occurs as part of play rather than as a separate evaluative activity. These quiz moments serve three roles within the activity: (i) they provide formative evidence, meaning information about the learner’s current vocabulary knowledge that helps tutors decide what to reinforce in follow-up practice, by indicating which target vocabulary items introduced in the narrative are already known versus still emerging, (ii) they function as lightweight attention checks by indicating whether learners are tracking the story and participating and (iii) in the presence of \textit{Maple}, quiz moments can be made more playful through contingent feedback to correct and incorrect selections, re-framing errors as supportive practice rather than a strict evaluation. At a system level, this design turns the storybook into a low-burden formative assessment instrument that is consistent with embedded, or \textit{``stealth''} assessment approaches \cite{shute2011stealth}, that aim to gather evidence of learning without disrupting engagement. Quiz interactions can be logged, including quiz answers and response time, and summarized at the end for tutors. This summary could be used by the tutor to plan what content is needed for the learner to revisit, supporting targeted reinforcement after the story activity.

\subsubsection{Support one-on-one attention in a triadic interaction (Guideline 5)}
A central outcome of the system design is support for sustained one-on-one engagement while preserving the tutor’s ability to intervene when needed. Consistent with our prior work \cite{fernandes2026codesigning}, \textit{Maple} is designed to operate primarily in a dyadic mode with the child for predictable and repeatable practice, while the tutor remains present and can shift the interaction into a triadic configuration at moments of confusion, clarification, or instructional emphasis. For example, during quiz segments, tutors can step in to explain target before returning the activity to the robot-led flow. This was inspired from the observations that were made during the shadowing sessions. The tutor intervention is implemented through a toggleable play and pause button in the interface, reflecting on an expert request for a mechanism that allows tutors to pause the story sequence and address uncertainty in the moment. The pause interaction creates ad-hoc transition points where tutors can re-explain vocabulary, or clarify instructions. The tutor can then resume the story when the learner is ready. This \textit{'tutor-in-the-loop'} design is consistent with evidence that effective one-on-one support is contingent and adaptive, with tutors diagnosing moment-to-moment understanding and adjusting scaffolding accordingly \cite{wood1976tutoring}. It is also aligned with accounts of robot-assisted learning in authentic settings, where teachers actively shape the interaction ecology by mediating breakdowns and validating children’s contributions, rather than merely supervising \cite{maijala2023teacherrole}.

\subsection{Emotional regulation using facial animations}

\textit{Maple} needed a facial display that could be controlled in real time so that affective cues could be contingent on interaction events while maintaining a supportive, non-evaluative tone. Prior work on robot-assisted learning for newcomer and refugee contexts has explored screen-based facial feedback using pre-scripted, progressions of facial states \cite{yang2024towards}. However, this approach is not suited for the current implementation of \textit{Maple}'s interaction flow, where facial behavior must be updated simultaneously with the robot's gestures during the autonomous sequences. We therefore adopted PyLips, which exposes API based control over expression parameters using facial Action Units, alongside reusable presets and a customization interface \cite{dennler2024pylips}. This choice supports event-contingent affective feedback, aligning with broader accounts that expressive behavior in sociable robots can help regulate and sustain social interaction and engagement when used as socially meaningful feedback \cite{breazeal2003emotion}.


Co-design discussions also highlighted the role of facial cues (used with PyLips) in supporting emotion regulation, particularly for young newcomer children. Experts emphasized that  \textit{``[...] the socio-emotional piece [is] important [...] part of what we do in our program is create that sense of community and reinforcement when everything is new for them''}. This observation shows that learners may benefit from reassurance during activities alongside language practice. To address this requirement, \textit{Maple} is able to render an expressive face on the phone-based display, enabling real-time affective feedback during the activity, using PyLips. Coupling facial display with embodied motion provides a mechanism for moment-to-moment \textit{``emotional adjustment''}, without shifting the interaction into explicit evaluation. This design also supports our positioning of \textit{Maple} as a peer-like companion \cite{fernandes2026codesigning}, where facial expressiveness functions as a social signal to sustain engagement and comfort while maintaining a supportive, non-evaluative tone throughout the interaction.

This socio-emotional role was discussed not only in terms of reinforcement, but also in terms of creating a \textit{``positive presence''} for learners who may have arrived in Canada only recently and are still adjusting to unfamiliar routines, spaces, and expectations. In this sense, facial expressiveness was not treated as a purely aesthetic feature, but as part of how \textit{Maple} could contribute to a more welcoming and encouraging atmosphere during activities. Rather than simply delivering instructions, the robot's animated face can help communicate warmth, enthusiasm, and gentle encouragement alongside story narration and guided tasks. Experts also suggested that this affective dimension may be especially valuable for children who are initially shy or hesitant to engage. As one tutor reflected, some learners may feel \textit{``[...] more comfortable engaging with the robot than with the tutor [...]''}. This insight reinforces our use of an expressive face for  facial feedback, which may help lower the interactional barrier for children who are not yet ready to respond confidently in front of an adult tutor. In that sense, facial animation supports not only engagement, but also the creation of a `softer` entry point into participation. The co-design discussions also positioned \textit{Maple} as a helper and companion rather than as an authority figure. This framing shaped how facial expressions were interpreted in the design. For example, a negative expression such as a frown was discussed not as a signal that the child had failed, but as a shared cue that the child and robot had encountered difficulty together, after which the robot could \textit{``suggest how to move forward''}, where such a framing may help preserve a non-judgmental tone. The L2 experts also mentioned that \textit{``Although Maple can't display or respond with in-depth emotion, these facial cues are crucial to having meaningful interactions with learners''}, supporting the idea that Maple's facial expressions could help \textit{``[...] maintain a positive reinforcement and encourage a welcoming environment [...]''}. We therefore designed \textit{Maple}'s facial animations to support encouragement, while preserving the tutor's primary facilitative role in the activity.

\subsection{TTS engine utilization and improvements}

The initial prototype used Kokoro TTS \footnote{\url{https://huggingface.co/spaces/hexgrad/Kokoro-TTS}} to generate speech for \textit{Maple} and for all story characters. Utterances are pre-generated as audio files and played through the user interface (for story characters and narrator voice) and the robot's phone (for \textit{Maple}'s voice). During internal testing, however, the resulting speech was perceived as less expressive and less natural than the voices typically encountered in commercial robotic systems. To obtain more consistent, higher-quality voice lines, we used Narakeet\footnote{\url{https://www.narakeet.com}} for script-to-audio generation for \textit{Maple} and most story characters, and supplemented this with limited human voice acting for selected roles using Alibaba Qwen3-TTS\footnote{\url{https://github.com/QwenLM/Qwen3-TTS}}. A recurring challenge for child-facing robots is that high-quality, child-like synthetic voices remain under-supported. Child-like TTS services may be difficult to be generated in part because large, representative datasets of children’s speech are scarce and harder to collect \cite{watts2008hmm}. As a result, models trained primarily on adult speech often do not match the pitch range and timbre that children expect from child-like voices, which can reduce perceived naturalness and social fit. We also explored multilingual speech support given the linguistic diversity among newcomer children in Canada. One option is to integrate a multilingual TTS model, e.g. Qwen3-TTS, which supports streaming speech generation across multiple languages. Another option investigated was using a multilingual language model, e.g. Aya\footnote{\url{https://cohere.com/research/aya}}, to translate utterances, prior to synthesis, which could support multilingual and multi-modal scaffolding. However, when generative models are used to produce new utterances at runtime, outputs could be unreliable \cite{matuszek2026reporting} by including hallucinations.

\section{Conclusion and Future Work}
This paper presented a co-design study with United for Literacy tutors and managers that informed \textit{Maple}, a table-top, peer-like socially assistive robot for supporting newcomer children’s English learning and cultural orientation in community literacy programs. Our findings support a tutor-mediated framing in which the robot provides low-pressure, structured practice while tutors remain the instructional and cultural anchor. We translated these insights into an integrated prototype with story-driven interaction, multi-modal scaffolding, and lightweight embedded check-ins. While these design choices were grounded in co-design and internal prototyping, they still require validation with children in practice. Future work will therefore focus on tutor playtesting and iterative child-centered case studies to examine usability, cue comprehension, perceived support, willingness to communicate, and any unintended effects in authentic sessions.

\section*{Acknowledgments}
Thank you to Nathan Dennler, Keith Rebello, Ali Yamini and Elaheh Sanoubari for their help and advice on the project.

\bibliography{citations}
\bibliographystyle{IEEEtran}

\end{document}